\documentclass[11pt,a4paper]{article}
\usepackage[hyperref]{emnlp2020}
\usepackage{times}
\usepackage{latexsym}

\usepackage{graphicx}
\usepackage{multirow}
\usepackage{microtype}
\usepackage{amssymb}
\usepackage{amsmath}
\usepackage{expex}
\usepackage{wrapfig}
\usepackage{cuted}

\usepackage{microtype}

\aclfinalcopy % Uncomment this line for the final submission
\title{Word Frequency Does Not Predict Grammatical Knowledge in Language Models}

\author{Charles Yu\thanks{ ~~These authors contributed equally.}, ~Ryan Sie\footnotemark[1], ~Nico Tedeschi, and Leon Bergen \\
    \texttt{\{cty002,rdsie,njtedeschi,lbergen\}@ucsd.edu} \\
  University of California, San Diego\\}

\date{}

\begin{document}
\maketitle
\begin{abstract}
Neural language models learn, to varying degrees of accuracy, the grammatical properties of natural languages. In this work, we investigate whether there are systematic sources of variation in the language models' accuracy.
 Focusing on subject-verb agreement and reflexive anaphora, we find that certain nouns are systematically understood better than others, an effect which is robust across grammatical tasks and different language models. Surprisingly, we find that across four orders of magnitude, corpus frequency is unrelated to a noun's performance on grammatical tasks. Finally, we find that a novel noun's grammatical properties can be few-shot learned from various types of training data. The results present a paradox: there should be less variation in grammatical performance than is actually observed.
\end{abstract}

\section{Introduction}

Neural language models \citep{howard2018universal,devlin2019bert,dai2019transformer, yang2019xlnet,radford2019language} have achieved success in both text prediction and downstream tasks such as question-answering, text classification, and natural language inference.
    %see \citep{otter2020survey}
    %and natural language inference \citep{conneau2017supervised, liu2018stochastic}.
The strong performance of these models raises scientific questions about the knowledge they have acquired, in particular, about the abstractness and generality of their linguistic representations. 

Previous work has investigated the linguistic representations of neural language models in several domains, and found varying evidence for how linguistically adequate these representations are \citep{lau2017grammaticality, marvin2018targeted, goldberg2019assessing, futrell2019neural}. This work has employed psycholinguistic methodology in order to elicit grammatical judgments from these models, inferring the models' underlying representations from the patterns of judgments.

In the current work, we focus on the variation in grammatical knowledge that potentially exists within a neural language model. Just as in human psycholinguistic tasks, previous work on neural LMs has observed variability in grammatical judgments between different sentences; not all violations of a grammatical constraint are judged to be equally bad.
It is not clear, however, whether there are systematic sources of variation in these judgments, and if so, what the sources are. 

We will focus on variation among lexical items, using English subject-verb agreement and reflexive anaphora as a case study.
We first ask whether language models learn the grammatical properties of some nouns more accurately than for others.
We do this by measuring the accuracy of language models when making grammatical judgments involving different nouns. We find systematic variation among nouns: nouns that perform well on one task or language model are more likely to perform well on other tasks or other language models.
We then consider possible sources of the observed variation between nouns, finding that the grammatical properties of nouns are paradoxically easy to learn; our results suggest that there should be much less variation than is actually observed.\footnote{All code and experimental materials are available at \url{https://github.com/CharlesYu2000/lm-variation}} 

\subsection*{Related work}

A number of other studies have investigated the linguistic representations of neural models, both language models specifically and networks trained using other objectives. \citet{linzen2016assessing, gulordava2018colorless, kuncoro2018lstms} probe the ability of LSTMs to learn hierarchical structures. \citet{warstadt2019neural} introduces a large-scale corpus of grammatical acceptability judgements, trains RNNs to predict these judgments, and concludes that the models outperform unsupervised baselines, but fall far short of human performance. \citet{lepori2020representations} finds that tree-based RNNs outperform sequential RNNs on number prediction tasks, but that fine-tuning on an artificially-generated augmentation set can bring the models closer to parity.

Other work has focused on probing whether neural language models have acquired adequate representations of specific linguistic phenomena. \citet{marvin2018targeted} and \citet{goldberg2019assessing} use a minimal pair methodology to assess the grammatical knowledge of RNNs and BERT, looking at subject-verb number agreement, reflexive anaphora, and negative polarity items. \citet{wilcox2018rnn} examines whether RNN language models exhibit \textit{wh}-licensing interactions on surprisal associated with gaps, concluding they can represent long-distance filler-gap dependencies and learn certain island constraints. \citet{futrell2019neural} studies whether neural language models show evidence for incremental syntactic state representations using psycholinguistic methodology. \citet{warstadt2019investigating} studies BERT's knowledge of NPI's, focusing on differences between tasks: boolean classification (e.g. \citealt{linzen2016assessing} and \citealt{warstadt2019neural}), minimal pair comparisons (e.g. \citealt{marvin2018targeted} and \citealt{wilcox2019structural}), and probing tasks (e.g. \citealt{giulianelli2018under}).
\section{Approach}
We use the minimal pair methodology of \citet{marvin2018targeted} in order to investigate the grammatical judgments of neural language models. Given a minimal pair of sentences, i.e. a pair that differ from each other in their acceptability due to a difference in just one grammatical property.
If the model understands the grammatical phenomenon being studied, it should assign higher probability to the grammatical sentence than to the ungrammatical sentence.

\subsection{Grammatical tasks}

Table \ref{tab:title} shows the 10 grammatical tasks \citep{marvin2018targeted} and the templates used for generating minimal pairs.
    %NICO: credit Marvin and Linzen for the templates here too?
The tasks fall into two general categories: subject-verb agreement (SVA) and reflexive anaphora (RA).
The first SVA task, SVA Simple, probes whether the model understands that subject number must agree with the number of third-person present verbs:
%% Example with parts
\pex[*=?*]<judgements>   %% Longest judgement mark
% first part
\a The cat walks.
% second part (with judgement}
\a \ljudge{*} The cat walk.
\xe
The other SVA tasks probe whether the models have more sophisticated representations of number agreement. For example, the SVA PP task measures whether the model is able to ignore distractors (``boys") which occur between the head of the subject and the verb:
\pex[*=?*]<judgements>   %% Longest judgement mark
% first part
\a The cat next to the boys jumps.
% second part (with judgement}
\a \ljudge{*} The cat next to the boys jump.
\xe
The object relative clause tasks probe whether the model accurately maintains the head's number in the presence of an embedded clause. \citet{marvin2018targeted} provide extensive discussion of the linguistic motivation for these tasks.

The RA tasks measure whether the language model understands the structural conditions on the binding of reflexive pronouns. The tasks make use of the following property of English reflexives: a reflexive pronoun needs to agree in number with its antecedent. The RA Sent.Comp task evaluates whether the model understands that reflexives must be in the same clause as their antecedents:
\pex[*=?*]<judgements>   %% Longest judgement mark
% first part
\a The lawyers said the defendant incriminated himself.
% second part (with judgement}
\a \ljudge{*} The lawyers said the defendant incriminated themselves.
\xe
The RA tasks involving object relative clauses evaluate whether the models understand that reflexive anaphora do not bind to the noun in an embedded clause but rather to the head noun.

\begin{table*}
\centering
\resizebox{0.8\paperwidth}{!}{%
\begin{tabular}{|l||l|}
    \hline
    \bf Task & \bf Template \\
    \hline\hline
    SVA Simple & The $\langle$TargetNoun$\rangle$ $\langle$Verb$\rangle$. \\
    SVA Subj.Rel.Clause & The $\langle$TargetNoun$\rangle$ that liked the $\langle$Noun$\rangle$ $\langle$Verb$\rangle$. \\
    SVA Sent.Comp. & The $\langle$Noun$\rangle$ said the $\langle$TargetNoun$\rangle$ $\langle$Verb$\rangle$. \\
    SVA PP & The $\langle$TargetNoun$\rangle$ next to the $\langle$Noun$\rangle$ $\langle$Verb$\rangle$. \\
    SVA Obj.Rel.Clause.That & The $\langle$TargetNoun$\rangle$ that the $\langle$Noun$\rangle$ liked $\langle$Verb$\rangle$. \\
    SVA Obj.Rel.Clause.NoThat & The $\langle$TargetNoun$\rangle$ the $\langle$Noun$\rangle$ liked $\langle$Verb$\rangle$. \\
    RA Simple & The $\langle$TargetNoun$\rangle$ $\langle$PastTransVerb$\rangle$ $\langle$himself/themselves$\rangle$. \\
    RA Sent.Comp. & The $\langle$NonGenderedNoun$\rangle$ said the $\langle$TargetNoun$\rangle$ $\langle$PastTransVerb$\rangle$ $\langle$himself/themselves$\rangle$. \\
    RA Obj.Rel.Clause.That & The $\langle$TargetNoun$\rangle$ that the $\langle$NonGenderedNoun$\rangle$ liked $\langle$PastTransVerb$\rangle$ $\langle$himself/themselves$\rangle$. \\
    RA Obj.Rel.Clause.NoThat & The $\langle$TargetNoun$\rangle$ the $\langle$NonGenderedNoun$\rangle$ liked $\langle$PastTransVerb$\rangle$ $\langle$himself/themselves$\rangle$. \\
    \hline
\end{tabular}}
\captionof{table}{Templates used for sentence generation. TargetNoun indicates the position of the target noun whose performance score is being calculated.} \label{tab:title}
\end{table*}

\subsection{Measuring the performance of a noun}

We use these tasks in order to measure how well the model understands the grammatical properties of a particular \emph{target noun}. Given a specific target noun, it is substituted as the TargetNoun in each of the task templates shown in Table \ref{tab:title}.
    %NICO: "as the $\langle$TargetNoun$\rangle$"
This gives a partially specified template. For example, substituting the target noun ``zombie" in the SVA Simple template results in:
\pex[*=?*]<judgements>   %% Longest judgement mark
The zombie $\langle$Verb$\rangle$.
\xe
Given each of these partially specified templates, 500 minimal pairs are randomly sampled by filling in the remaining lexical items. Finally, the model's grammatical judgments on the 500 minimal pairs are computed (by taking the difference in scores between the grammatical and ungrammatical variants) and averaged, resulting in a task performance score for the noun.
    %NICO: ""

\subsection{Limitations}

These analyses are limited in several respects. First, only two grammatical tasks are used. By using a wider range of tasks, it will be possible to investigate a larger set of grammatical phenomena outside of number agreement.

Second, while the study focuses on the grammatical information carried by nouns, other lexical types such as verbs are likely to carry this information as well. Future work can determine whether the approach generalizes to verbs and other lexical types. 

Finally, while the study uses acceptability judgments in order to determine the models' grammatical knowledge, other probing tasks exist and may produce different results \citep{warstadt2019investigating}. We use acceptability judgments because, to the best of our knowledge, feature probing has not been extensively studied for GPT-2 or Transformer-XL. Different probing architectures may produce different results for these models. It would be desirable to understand the robustness of the current results to the choice of experimental readout. 

\section{Methods}
In this section we describe the process of calculating a target noun's task performance score in more detail.

\subsection{Sentence generation}\label{sec:sentence-generation}

Using WordNet \citep{fellbaum1998wordnet} and VerbNet \citep{schuler2005verbnet}, we compiled a list of lexical items as shown in Table \ref{tab:word-count}. The target nouns were drawn from the Noun list, which consisted of animate nouns. Only nouns with distinct singular and plural forms were included. All verbs in the Verb set have an intransitive reading.
For each pair of task template and target noun, 500 sentences were randomly sampled by choosing lexical items from the appropriate word lists. 

For each sampled sentence, 2*2 or 2*2*2 versions were generated (depending on the template). These versions varied the grammaticality of the sentence and the plurality of the target noun and any distractor nouns. For example, for the SVA Simple task, 2*2 versions are generated for every sampled sentence:
\pex \label{ex:variants}
\a Singular-Grammatical: The horse walks.
\a Singular-Ungrammatical: \ljudge{*} The horse walk.
\a Plural-Grammatical: The horses walk.
\a Plural-Ungrammatical: \ljudge{*} The horses walks.
\xe

\subsection{Models}
Our experiments use three models, Transformer-XL \citep{dai2019transformer}, GPT-2 \citep{radford2019language}, and BERT \citep{devlin2019bert}.
We use the Hugging Face implementations \citep{Wolf2019HuggingFacesTS} with the pre-trained models \textit{transfo-xl-wt103}, which is trained on the WikiText-103 dataset, \textit{gpt2-xl}, which is trained on the WebText dataset, and \textit{bert-base-uncased}, which is trained on BookCorpus and English Wikipedia.
    %NICO: maybe use \texttt (monospaced) instead of \textit for the models

\begin{table}
\centering
\resizebox{\columnwidth}{!}{%
\begin{tabular}{|c||c|c|c|}
    \hline
    \bf Set Name &  Transformer-XL & GPT-2 & BERT \\
    \hline\hline
    Noun & 916  & 723 & 704  \\
    Verb & 615 & 228 & 406 \\
    NonGenderedNoun & 870  & 679  & 663 \\
    PastTransVerb & 1298  & 1034 & 1298 \\
    \hline
\end{tabular}}
\captionof{table}{Size of word sets for each model.} \label{tab:word-count}
\end{table}

\subsection{Sentence scoring}
We now describe how a score was calculated for a particular sampled sentence. 
For each of the sentence variants (e.g. Example \ref{ex:variants}), the model computes a score. In the case of Transformer-XL and GPT-2, this score is simply the the log probability of the string. For example, for Transformer-XL:
\begin{equation}
    \text{Score}_{\text{string}}(s) = \log P_{\text{TXL}}(s)
\end{equation}
where $P_{\text{TXL}}$ is the Transformer-XL language model probability distribution. 

For BERT, given its masked language model architecture, we follow the approach of \citet{goldberg2019assessing}. For the SVA tasks, we compute the log conditional probability of the verb whose number must agree with the target noun. For the RA tasks, we compute the log conditional probability of the reflexive pronoun. Both conditional probabilities are computed conditional on the left and right contexts.

Given the scores for a sentence's variants, we compute an overall score for the sentence, which captures how much the model prefers the grammatical variants to the ungrammatical variants. For each sampled sentence $S$, there are either 2 or 4 minimal pairs among its variants. In Example \ref{ex:variants}, a. and b. is a minimal pair, and c. and d. is a minimal pair. Letting $s_a,...,s_d$ denote these variants, the overall score for the sentence is given by:
\begin{align*}
    \text{Score}_{\text{sent}}(S) = \frac{1}{2}(&\text{Score}_{\text{string}}(s_a)-\text{Score}_{\text{string}}(s_b) \\ + &\text{Score}_{\text{string}}(s_c)-\text{Score}_{\text{string}}(s_d))
\end{align*}
The formula when there are four minimal pairs is similar.

\subsection{Noun scoring}\label{sec:noun-scoring}
We next compute an overall score for the target noun.
As described in Section \ref{sec:sentence-generation}, for a specific target noun $n$ and task, we sample 500 sentences $S_1,...,S_{500}$. 
The noun's score for this task is then given by:
\begin{equation}
    \text{Score}_{\text{noun}}(n) = \frac{1}{500} \sum_{i=1}^{500}\text{Score}_{\text{sent}}(S_i)
\end{equation}

\subsection{Word filtering and tokenization}

Words were removed from a particular model if either their singular or plural form was tokenized to \emph{unk}, or if their singular and plural forms were assigned different numbers of tokens.\footnote{The latter constraint was used in order to simplify batching.} For BERT, words in the Verb set were removed if they were assigned more than one token, as BERT does not model the joint distribution over multiple masked tokens.

For Transformer-XL, we add a padding text\footnote{\url{https://tinyurl.com/y9kjuj5q}} and a start-of-sentence-token ($\langle$SOS$\rangle$) to the beginning of the sentence and an end-of-sentence token ($\langle$EOS$\rangle$) to the end of the sentence. 
For GPT-2, we make no modifications to the generated sentence (although prefix spaces are added to the strings for tokenization purposes). 
For BERT, since it is a masked language model, we replace the Verb (for SVA) or reflexive pronoun (for RA) with a [MASK] token after tokenization. 
Thus, each sentence will have a single mask token corresponding to the word that should agree with the target noun. 
\section{Results}\label{sec:results}

\subsection{Noun performance is correlated across tasks}\label{sec:task-correlation}

We first examine how each noun's performance varies across the grammatical tasks. For each noun-task pair, we measure the average performance of the noun on that task, as described above. This gives 10 features per noun, corresponding to the 10 grammatical tasks. 

\begin{figure}
    \includegraphics[width=\columnwidth, keepaspectratio]{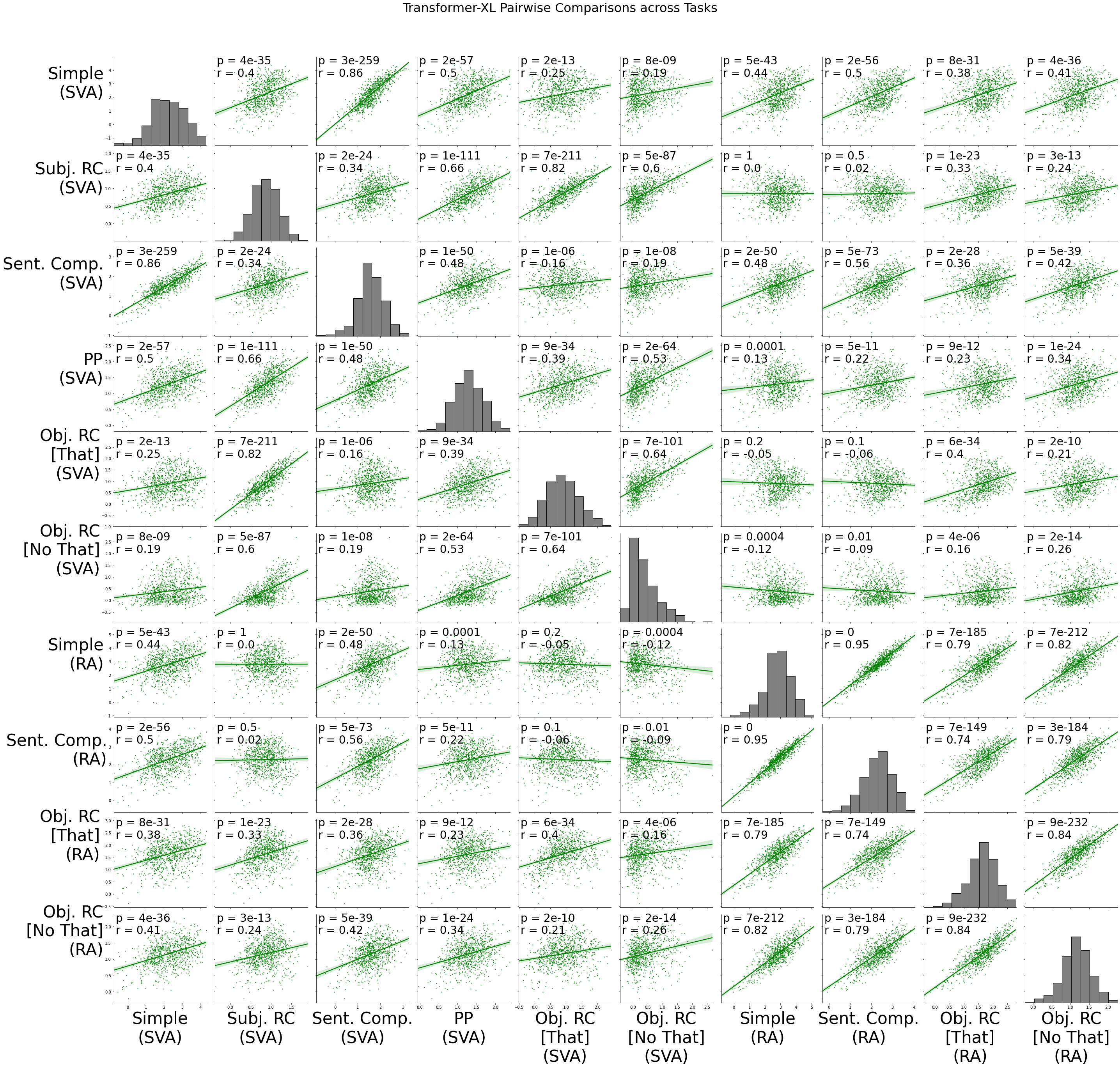}
    \caption{Pairwise comparisons between tasks with Transformer-XL. Rows and columns represent tasks, and one point represents a single noun's performance on a pair of tasks. The four tasks on the lower right, with strongest correlations, all involve reflexive anaphora.}
    \label{fig:txl_full_pairplot}
\end{figure}

Figure \ref{fig:txl_full_pairplot} shows the pairwise comparisons between performance on the different tasks for Transformer-XL. Results for BERT and GPT-2 are similar and are shown in the appendix. The figure shows that performance is correlated across the tasks; for many pairs of tasks, nouns which have higher performance on one task are likely to have higher performance on the other.

Using principal component analysis, we found that a single principal component explains 47\% of task variance for Transformer-XL, and two principal components explain 73\%. Results are similar for BERT and GPT-2, and are shown in the appendix. The first PC primarily measures performance on the four reflexive anaphora tasks, while the second PC measures performance on the subject-verb agreement across relative clause tasks. This suggests that there is a dimension that characterizes whether the model understands how reflexive binding constraints operate for a noun, and a dimension for whether the model understands subject-verb agreement for the noun. Note that Figure \ref{fig:txl_full_pairplot} additionally demonstrates correlations between the reflexive tasks and the subject-verb agreement tasks.

These results provide evidence that language models' variation in performance on the grammatical tasks is, in part, explained by properties of the nouns which are stable across tasks. The models understand number agreement better for some nouns, and worse for others.

\subsection{Noun performance is correlated across models}\label{sec:model-correlation}

We next investigate whether nouns exhibit stable behavior across different neural language models. For each pair of the three language models, we measured how well a noun's task performance in one language model predicted its task performance in the other language model. 

\begin{figure*}[t]
    \centering
    \includegraphics[width=0.8\paperwidth, keepaspectratio]{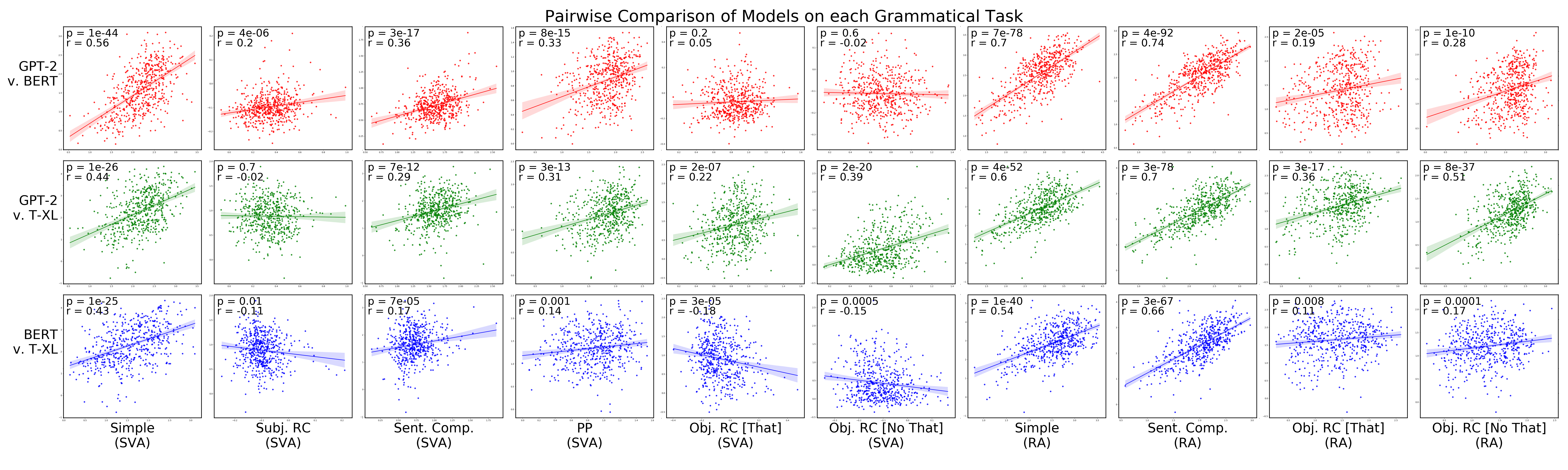}
    \caption{Pairwise comparisons between GPT-2, Transformer-XL, and BERT on the 10 grammatical tasks. Each row corresponds to a pair of language models, and each column is a single task. One point represents the performance of a noun on a single task.}
    \label{fig:model-correlations}
\end{figure*}

Figure \ref{fig:model-correlations} shows comparisons between pairs of language models on the 10 grammatical tasks. Of the 30 comparisons, 24 show significant positive correlations between the pairs of language models. 22 of the correlations remain significant after Bonferroni correction.  

GPT-2 and Transformer-XL show the strongest correlation in performance. It is possible that this is due to methodological differences between the task setup for GPT-2 and Transformer-XL compared to BERT: GPT-2 and Transformer-XL are performing a language modeling task in which the probability of a full sentence is queried, while BERT performs masked language modeling on a single target word. The difference may also be due to corresponding training differences between BERT and the autoregressive language models.

The results provide evidence that nouns exhibit stable task performance across language models. The source of the correlation across language models must come from features of the training data. Properties of the natural text distribution of nouns lead some of these nouns to be better understood than others.

\subsection{Effect of frequency on task performance}

\begin{figure*}
    \centering
    \includegraphics[width=0.8\paperwidth]{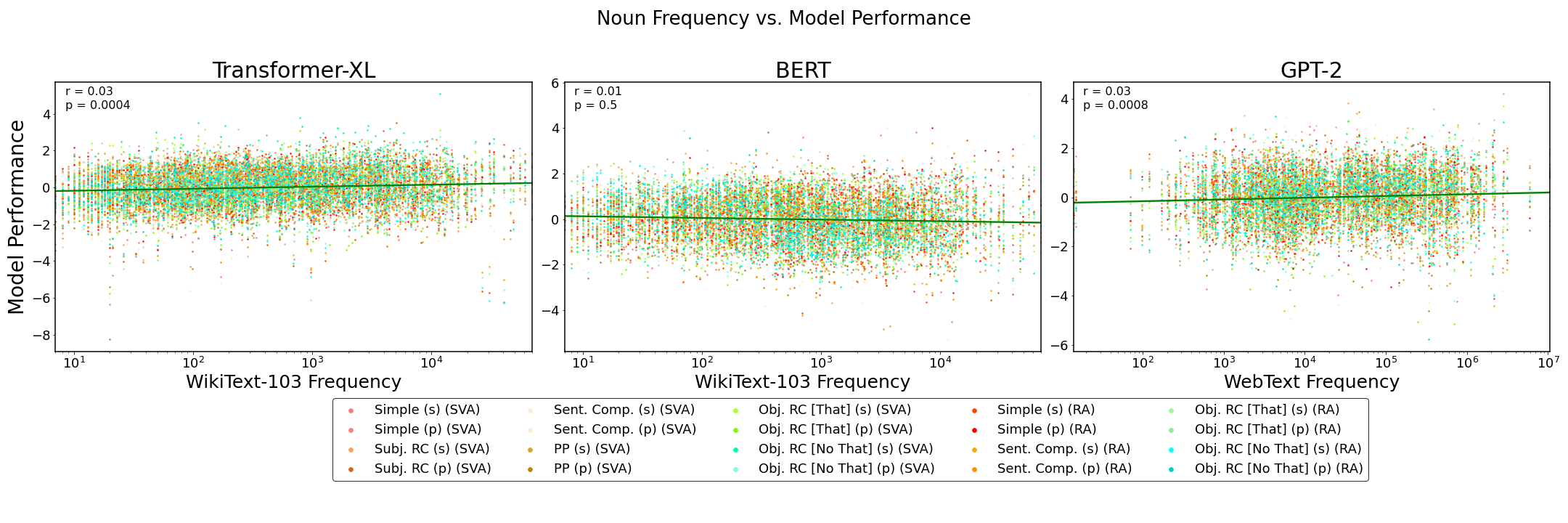}
    \caption{Relationship between corpus frequency and task performance for Transformer-XL, BERT, and GPT-2. Performance scores are z-normalized. Colors indicate the ten grammatical tasks and singular/plural form of the noun (\emph{s} indicates singular, \emph{p} indicates plural). Each point represents task performance for a single noun. }
    \label{fig:model-frequency}
\end{figure*}

In Sections \ref{sec:task-correlation} and \ref{sec:model-correlation}, we found evidence that nouns exhibit stable performance across different grammatical tasks and language models. One obvious explanation of these results is that nouns vary in their frequency in natural text, and language models learn more accurate grammatical representations for more frequent nouns.

In order to investigate this, we measured the frequency of each noun in two corpora: WikiText-103, a 103 million token subset of Wikipedia, which was used for training Transformer-XL; and Open WebText \citep{Gokaslan2019OpenWeb}, an open-source implementation of the web corpus used to train GPT-2.\footnote{BERT was trained on a mix of Wikipedia text and BookCorpus. Because, as of this writing, BookCorpus is no longer distributed, WikiText-103 was used as a proxy for BERT training frequencies.} Word frequencies were measured separately for singular and plural noun forms. Figure \ref{fig:model-frequency} shows the relationship between frequency and task performance on each of the ten grammatical tasks. The appendix shows the results broken down by task type.

The results show no clear relationship between noun frequency and task performance. Frequency explains no more than $0.1\%$ of the variation in performance. This holds true over more than four orders of magnitude in frequency. This provides evidence that 1) differences in corpus frequency do not explain the systematic differences observed between nouns, and 2) relatively few observations suffice for transformer language models to learn correct number agreement behavior for a noun. In the next section, we investigate this finding further.

\section{Few-shot learning for novel lexical items}

The results in the previous section provide evidence that nouns systematically vary in their performance on grammatical tasks; some nouns perform better than others across tasks and language models. However, this variation is not explained by frequency of occurrence in natural text. Nouns that occur on the order of 100 times in a corpus do not have systematically worse performance than nouns that occur $10^6$ times. 

The results raise a question: if frequency does not influence how well a noun is understood, what does? If low frequency nouns are understood as well as higher frequency nouns, then this suggests that language models few-shot learn the grammatical properties of nouns. We suggest that by studying what makes a noun learnable in a few-shot setting, it may be psosible to better understand the sources of the observed variation.

We use a few-shot learning paradigm, introducing a new lexical item into the vocabulary of the language model, either ``wug" (intended as a new singular noun), or ``wuz" (intended as a plural). We then fine-tune the language model using several example sentences containing this word. 
    %NICO: something more specific than ``several"
Note that this paradigm is distinct from nearly all of the few-shot learning experiments performed in \citet{radford2019language,brown2020language}, which operate on a known vocabulary.\footnote{\citet{brown2020language} perform several experiments on novel vocabulary items.}

\subsection{Learning agreement from syntactic data}\label{sec:syntax-fine-tuning}

We first look at whether training data containing explicit syntactic markers of number agreement is sufficient for few-shot learning. Table \ref{tab:syntax-training} describes the types of training data we examine. The three types of training data use different syntactic markers of plurality to indicate whether the new noun is singular or plural. 

The language models are fine-tuned with 5 sentences drawn from a single training data type. GPT-2 was fine-tuned for 2 epochs, and BERT was fine-tuned for 4 epochs.\footnote{Prior to more systematic experiments, we informally optimized the number of fine-tuning epochs.} Transformer-XL was not used for the fine-tuning experiments, due to issues with introducing new vocabulary items given Transformer-XL's adaptive weight embedding. 

After fine-tuning, each model was evaluated on the 10 grammatical tasks in Table \ref{tab:title}. For each grammatical task, 500 sentences were sampled from the task template, and a performance score was calculated by averaging scores of the samples, as described in Section \ref{sec:noun-scoring}. 

\begin{figure}[]
    \includegraphics[width=1\columnwidth]{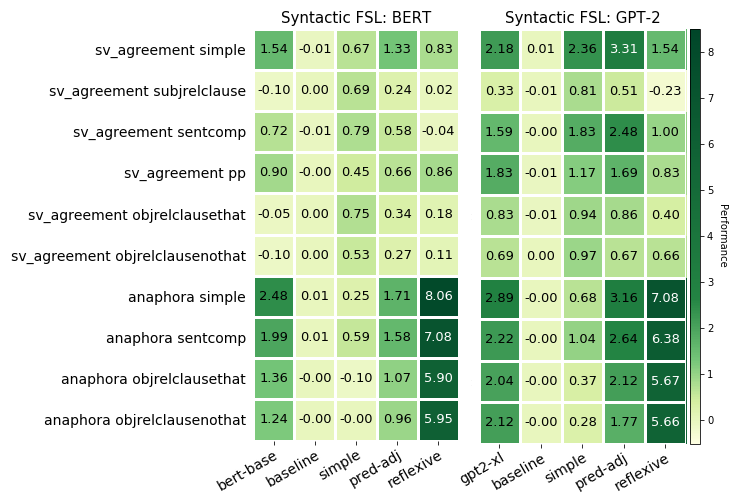}
    \caption{Few-shot learning from syntactic examples (averaging over plural and singular results). Columns show different types of training data, and rows show the 10 grammatical tasks. The bert-base and gpt2-xl columns indicate model performance on known lexical items, i.e. summarizing results from Section \ref{sec:results}. The baseline columns indicate performance of non-fine-tuned models on the novel wug/wuz lexical items. Scores are differences of log-probabilities between grammatical and ungrammatical. The 95\% confidence interval around each point estimate is always smaller than $\displaystyle \pm 0.25$.}
      \label{fig:few-shot-syntax}
\end{figure}

Figure \ref{fig:few-shot-syntax} shows results for fine-tuning on the three types of syntactic data. Compared to model performance on real lexical items (shown in the leftmost column), both BERT and GPT-2 achieve qualitatively similar performance given the Pred-adj and Reflexive training data, but worse performance given the Simple training data. Performance is weakest on subject-verb agreement (SV-agreement) tasks involving relative clauses. When trained on data containing reflexive anaphora, both models achieve notably higher performance on the grammatical tasks involving reflexive anaphora.

The results provide evidence that small amounts of syntactic training data support learning the agreement properties of novel nouns. They also provide evidence of heterogeneity among different types of training data. Training from bare present tense verbs is least effective, and training from sentences containing reflexives leads to improved performance on tasks which require understanding of the conditions on reflexive binding.

\begin{table}
\centering
\resizebox{\columnwidth}{!}{%
\begin{tabular}{|l||l|}
    \hline
    \bf Training data type & \bf Template \\
    \hline\hline
    Simple &  The wug/wuz $\langle$PresentTenseVerb$\rangle$. \\
    Pred-adj &  The wug/wuz is/are $\langle$Adj$\rangle$.  \\
    Reflexive & The wug/wuz $\langle$Verb$\rangle$ himself/themselves.  \\
    \hline
\end{tabular}}
\captionof{table}{The three types of training data used for syntactic fine-tuning.} \label{tab:syntax-training}
\end{table}

\subsection{Learning agreement from semantic data}

We next examine whether purely semantic indicators of plurality are sufficient for learning a noun's number agreement properties. 
We look at several types of constructions which provide information about the plurality of a noun, but using predicates with past tense verbs that don't inflect for number so that there is no grammatical number agreement. %NICO
In particular, we note the different possible readings with reference to the distributive and collective distinction described in the semantics literature \citep{lonning1997plurals, lasersohn2011mass, champollion2015distributivity}. %NICO
For documentation of predicates that require a collective NP subject, see \citet{levin1993english}. %NICO

We use the fine-tuning method from Section \ref{sec:syntax-fine-tuning}.

\subsubsection*{Singular constructions}
In order to induce singular noun interpretations, we use the singular-biased constructions shown at the top of Table \ref{tab:semantic-training}.
For example, if a wug worked all alone or came unaccompanied, it is likely that ``wug" is both semantically and grammatically singular. 
However, these constructions do not gramatically require the head noun to be singular: they are compatible with distributive readings where the predicate individually applies to members of a group (e.g. ``the lawyers worked all alone" means \textit{each} lawyer worked alone). %NICO

\begin{table}
    \centering
    \resizebox{\columnwidth}{!}{%
	\begin{tabular}{|c|l|l|}
			\hline
			& \multicolumn{1}{c|}{\bf Training data type} & \multicolumn{1}{c|}{\bf Example} \\ \hline
			\parbox[t]{2mm}{\multirow{4}{*}{\rotatebox[origin=c]{90}{\bf Singular}}} 
			%& sole & The wug became the sole owner. \\ \cline{2-3}
			%& lone & The wug became the lone survivor. \\ \cline{2-3}
			%&only & The wug became the only student. \\ \cline{2-3}
			&all-alone & The wug worked all alone. \\ \cline{2-3}
			&unaccompanied & The wug came unaccompanied. \\ \cline{2-3}
			&separated-entire & The wug became separated from the entire group. \\ \cline{2-3}
			&personally & The wug personally thanked me. \\ \hline\hline
			\parbox[t]{2mm}{\multirow{6}{*}{\rotatebox[origin=c]{90}{~~~~~\bf Plural}}} 
			&unison & The wuz nodded in unison. \\ \cline{2-3}
			&together & The wuz ate together. \\ \cline{2-3}
			&simultaneously & The wuz jumped simultaneously. \\ \cline{2-3}
			&outnumbered & The wuz outnumbered the cats. \\ \cline{2-3}
			&constituted & The wuz constituted a majority of the team. \\ \cline{2-3}
			&gathered & The wuz gathered quietly. \\ \hline
	\end{tabular}}
	\captionof{table}{Types of training data used for semantic fine-tuning.} \label{tab:semantic-training}
	
\end{table}

\begin{figure}[]
\hspace*{-0.5cm}  
    \includegraphics[width=1.1\columnwidth]{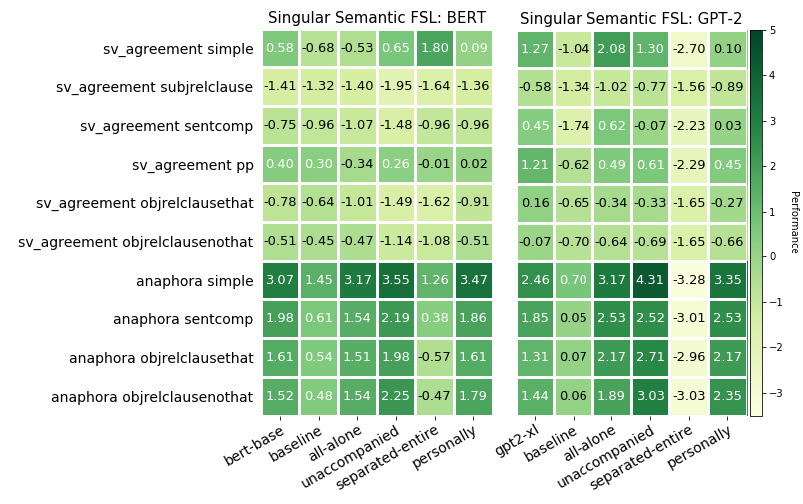}
    \caption{Few-shot learning from singular semantic examples. The bert-base and gpt2-xl columns indicate model performance on known lexical items, i.e. summarizing results from Section \ref{sec:results}. The baseline columns indicate performance of non-fine-tuned models on the novel wug token.}
      \label{fig:few-shot-semantic-singular}
\end{figure}

BERT and GPT-2 were fine-tuned on 5 examples of each of the singular constructions. Figure \ref{fig:few-shot-semantic-singular} shows the results. None of the constructions consistently induced correct performance on the grammatical tasks across both models. Three of the constructions --- \emph{all-alone}, \emph{unaccompanied}, and \emph{personally} --- led to strong performance on the reflexive anaphora tasks (stronger than the average performance calculated in Section \ref{sec:results}). The \emph{separated-entire} construction consistently decreased performance on the tasks relative to baseline.

\subsubsection*{Plural constructions}

In order to provide the models with data indicating that a novel noun is plural, we use constructions which force either collective or distributive readings. For example, in Table \ref{tab:semantic-training}, if the wuz constituted the majority of the team, then the word ``wuz" must be semantically plural. The construction \emph{constituted a majority} is collective because it must apply to the group as a whole:
\pex The doctors constituted a majority of the team.
\a *Distributive reading: each of the doctors constituted a majority.
\a Collective reading: the doctors as a group constituted a majority.
\xe
While the argument of a collective predicate must be semantically plural, it is not necessarily grammatically plural. For example, the singular ``the group" could constitute the majority of the team.

Three of the constructions in Table \ref{tab:semantic-training} are collective: \emph{outnumbered}, \emph{constituted}, and \emph{gathered}. The other three are distributive phrasal predicates, which force distributive readings:
\pex The architects nodded in unison.
\a Distributive reading: each of the architects nodded.
\a *Collective reading: the group of architects itself nodded.
\xe

\begin{figure}[]
    \hspace*{-0.7cm}  
    \includegraphics[width=1.15\columnwidth]{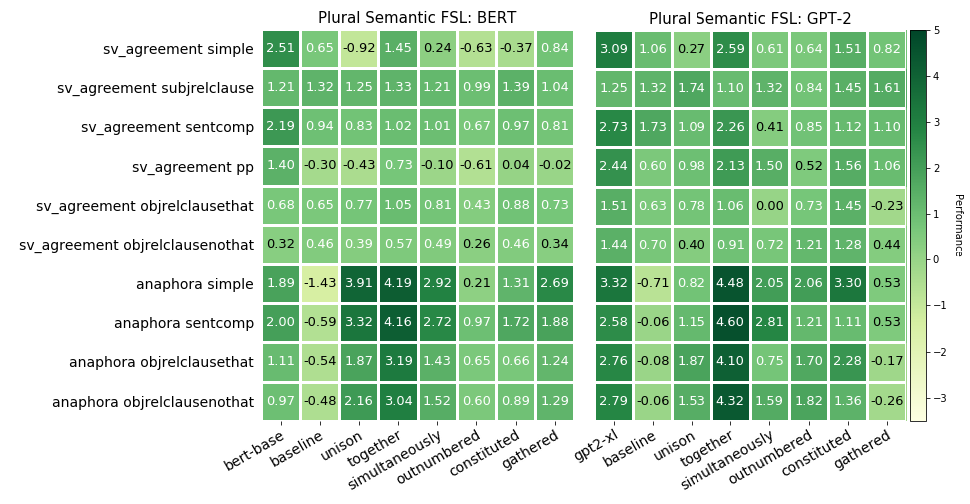}
    \caption{Few-shot learning from plural semantic examples. The baseline columns indicate performance of non-fine-tuned models on the novel wuz token.}
      \label{fig:few-shot-semantic-plural}
\end{figure}

Figure \ref{fig:few-shot-semantic-plural} shows the plural learning results. The 6 types of training data perform comparably on the subject-verb agreement tasks (and similar to the baseline model, which represents performance prior to fine-tuning). The three distributive phrasal constructions perform better on the reflexive anaphora tasks than the three collective constructions, though all constructions improve relative to the baseline.

\section{Discussion}

We have investigated the sources of variation in neural language models' grammatical judgments. We found that there are systematic differences between nouns: when a language model exhibits knowledge of a noun's grammatical properties in one task, it is more likely to do so in other tasks. Moreover, when one language model exhibits this knowledge, other language models are more likely to as well. The study found two latent dimensions of variation between nouns: one corresponding to how well the models understood its behavior with reflexive pronouns, and the other corresponding to subject-verb agreement.

Subsequent analyses demonstrate a pair of empirical phenomena:
\begin{enumerate}
\item It is relatively easy to learn the number agreement properties of a noun. The models learn the agreement properties of a novel noun from just a few samples, and the data supporting few-shot learning appears to be densely distributed; nearly all types of syntactic and semantic data examined lead to improvements on the reflexive pronoun or subject-verb agreement tasks. 
\item Nouns that occur more frequently during training are not learned more accurately. Many nouns that occur with high frequency are not learned accurately.
\end{enumerate}

These results suggest that nouns should vary less in their grammatical performance than is actually observed; the study finds \emph{excess variation} in grammatical performance. If number agreement can be correctly learned from a few samples (FSL samples), then one would expect model performance to either a) improve with more data, as more FSL samples are observed, or b) improve with more data up to some threshold, and then asymptote after learning has saturated. In either case, for high frequency nouns, a sufficient number of FSL samples should be observed for these nouns to be learned very accurately.

A potential explanation of the results is that they are caused by catastrophic forgetting \citep{ratcliff1990connectionist,french1999catastrophic}: although a sufficient number of FSL samples are observed for a noun, these samples are forgotten during training, causing the performance of the noun to degrade. This explanation is implausible. If catastrophic forgetting is occurring, then the problem should be more severe for infrequent nouns than for frequent nouns, as the interval between training samples will be longer for infrequent nouns. This would predict better performance for frequent nouns.

\section*{Acknowledgements}

 We thank the Google Cloud Platform research program for support. The Titan V used for this research was donated by the NVIDIA Corporation.

\bibliography{references}
\bibliographystyle{acl_natbib}
\appendix
\section{Further analyses}
\begin{figure*}[h!]
\centering
\resizebox{0.7\paperwidth}{!}{%
    \includegraphics[width=\paperwidth,keepaspectratio]{"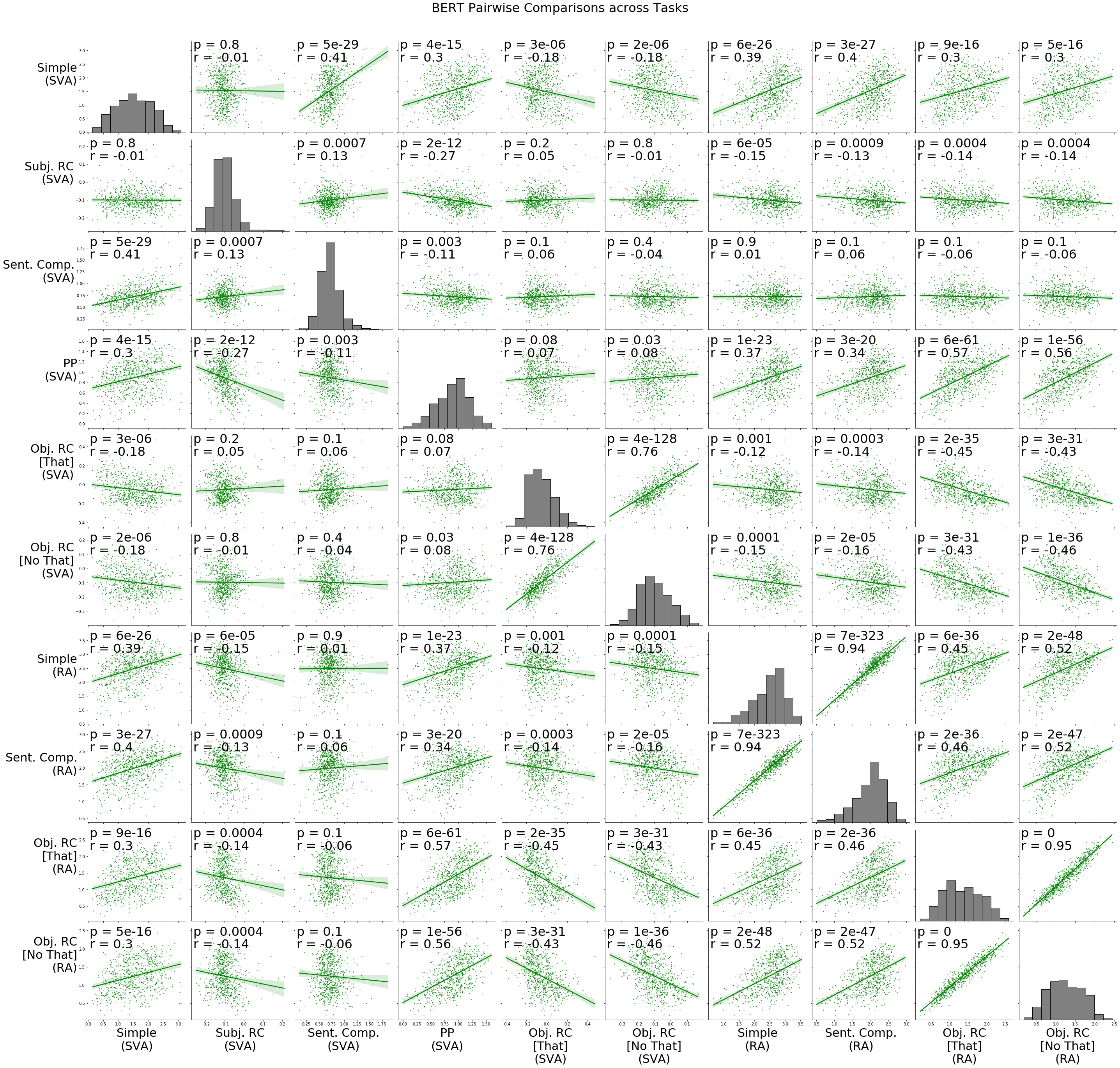"}
}
    \caption{BERT: Pairwise comparisons between tasks. }
    \label{fig:bert_pairplot}
\end{figure*}

\begin{figure*}[h!]
\centering
\resizebox{0.7\paperwidth}{!}{%
    \includegraphics[width=\paperwidth,keepaspectratio]{"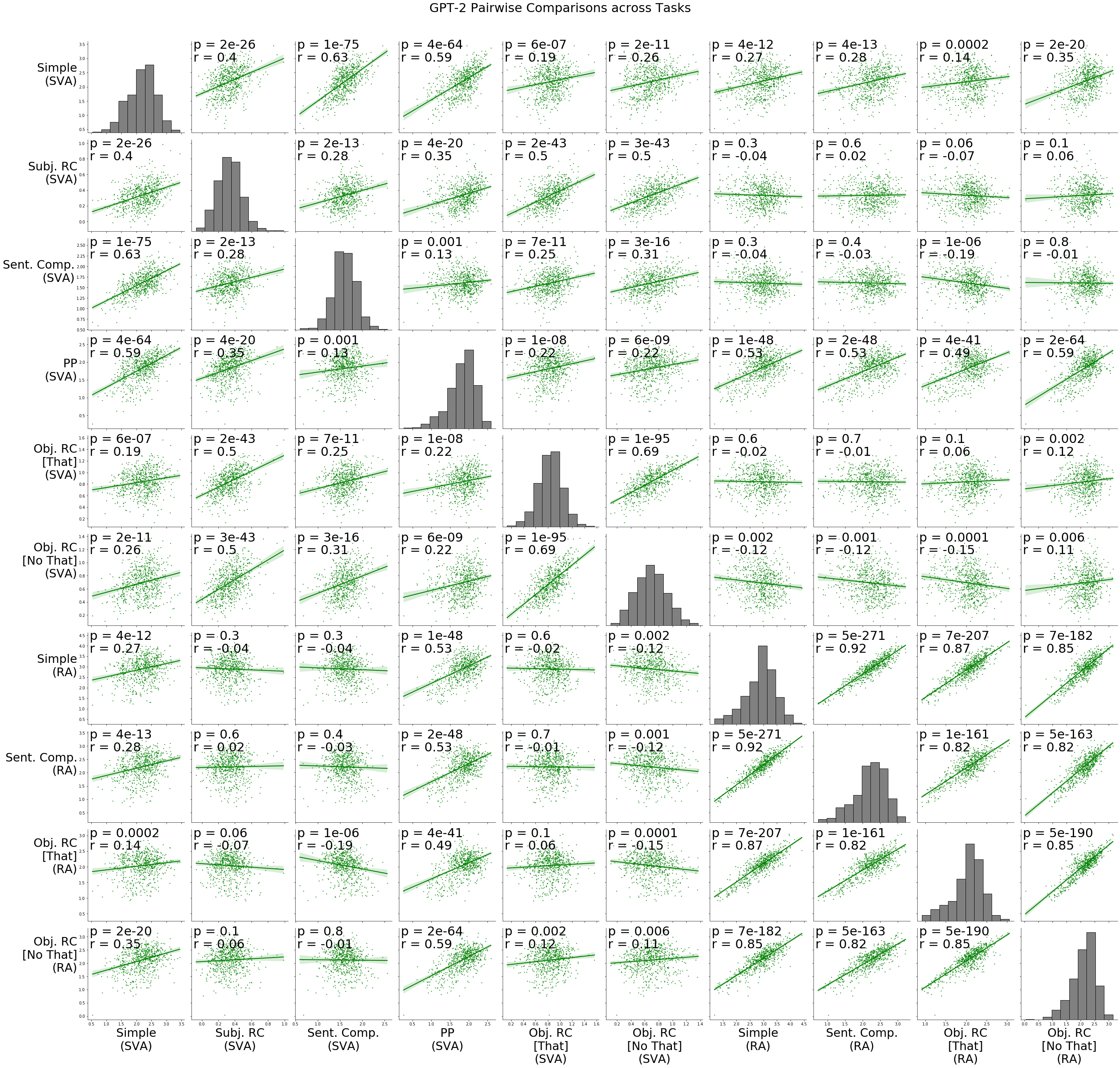"}
}
    \caption{GPT-2: Pairwise comparisons between tasks.}
    \label{fig:gpt_pairplot}
\end{figure*}

\begin{figure*}[h!]
\resizebox{0.7\paperwidth}{!}{%
    \includegraphics[width=\paperwidth,keepaspectratio]{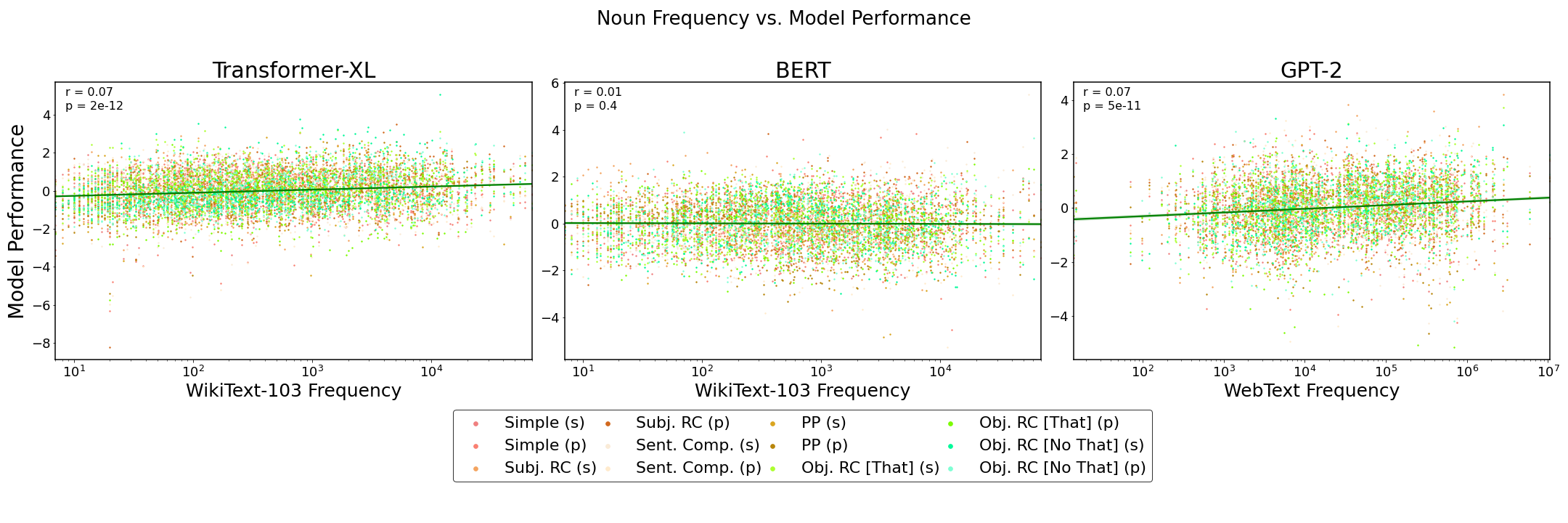}
}
    \caption{Noun Frequency vs. Model Performance on Subject-Verb Agreement Tasks}
    \label{fig:freq_sva}
\end{figure*}

\begin{figure*}[h!]
\resizebox{0.7\paperwidth}{!}{%
    \includegraphics[width=\paperwidth,keepaspectratio]{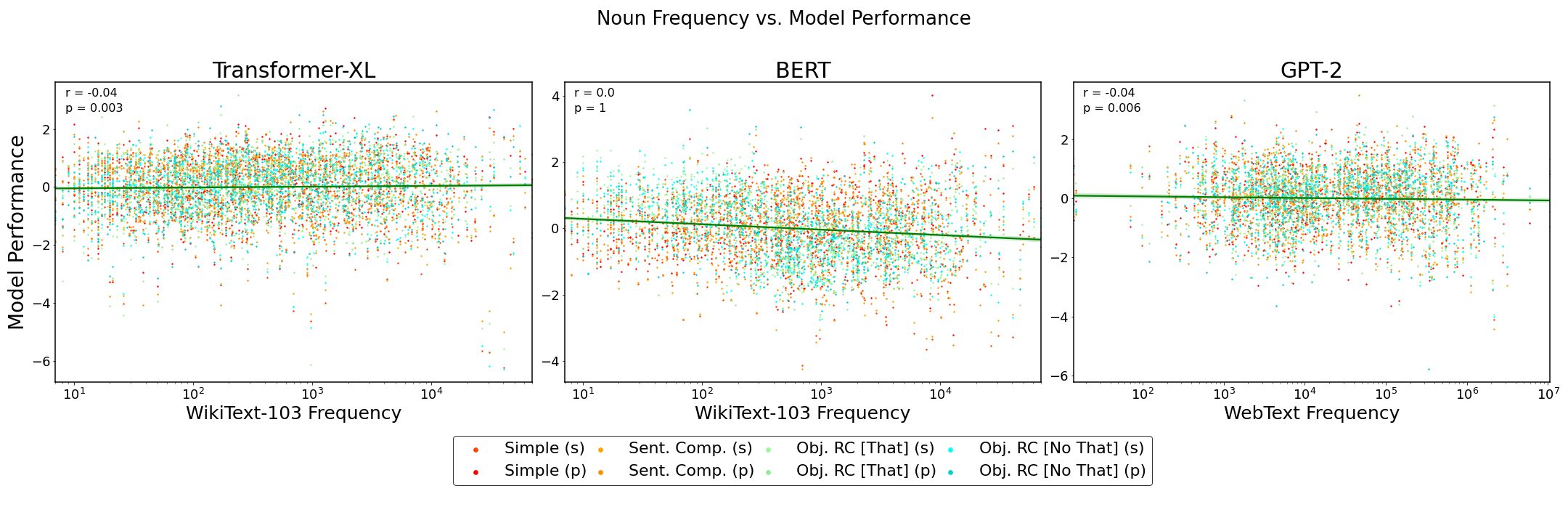}
}
    \caption{Noun Frequency vs. Model Performance on Reflexive Anaphora Tasks}
    \label{fig:freq_ra}
\end{figure*}

This section contains several additional analyses: principal components for task performance among the three language models (Tables \ref{tab:pca_variances}-\ref{tab:pca_contributions_bert}); pairwise comparison between task performance for BERT and GPT-2 (Figures \ref{fig:bert_pairplot} and \ref{fig:gpt_pairplot}); and more fine-grained comparisons between word frequency and model performance (Figures \ref{fig:freq_sva} and \ref{fig:freq_ra}).

\begin{table*}[h!]
    \centering
    % \resizebox{0.8\textwidth}{!}{%
	\begin{tabular}{|c||c|c|c|}
			\hline
			    PC Number & Transformer-XL & BERT & GPT-2 \\
			    \hline\hline
			    1 & 0.4663 & 0.3865 & 0.4146 \\
			    2 & 0.7299 & 0.5619 & 0.6873 \\
			    3 & 0.8511 & 0.7073 & 0.8059 \\
			    4 & 0.9083 & 0.8034 & 0.8720 \\
			    5 & 0.9499 & 0.8919 & 0.9175 \\
                % 6 & 0.96510187 & 0.941521736 & 0.949228816 \\
                % 7 & 0.977117721 & 0.967615448 & 0.969288059 \\
                % 8 & 0.988467316 & 0.990502723 & 0.985162928 \\
                % 9 & 0.995964251 & 0.996153056 & 0.994004225 \\
                % 10 & 1 & 1 & 1 \\
			\hline
	\end{tabular}
% 	} % resizebox
	\caption{Cumulative proportion of variance explained by the top (of 10) PCs for each model as detailed in Section 4.1.} \label{tab:pca_variances}
\end{table*}

% pca table - contributions - txl

\begin{table*}[h!]
\centering
\resizebox{0.8\paperwidth}{!}{%
\begin{tabular}{|c||l|l|l|}
		\hline
		    Contributor by Rank & PC 1 & PC 2 & PC 3 \\
		    \hline\hline
            1 & RA ObjRelClauseNoThat - 0.386980 & SV SubjRelClause - 0.449504 & SV SentComp - 0.534710 \\
            2 & RA ObjRelClauseThat - 0.376354 & SV ObjRelClauseNoThat - 0.442445 & SV Simple - 0.516031 \\
            3 & RA SentComp - 0.359096 & SV ObjRelClauseThat - 0.439015 & RA ObjRelClauseThat - 0.402452 \\
            4 & RA Simple - 0.347978 & RA Simple - 0.376192 & SV ObjRelClauseThat - 0.312466 \\
            % 5 & SV Simple - 0.344456 & RA SentComp - 0.352908 & RA ObjRelClauseNoThat - 0.292776 \\
            % 6 & SV SentComp - 0.342392 & SV PP - 0.303320 & SV PP - 0.236013 \\
            % 7 & SV PP - 0.280799 & RA ObjRelClauseNoThat - 0.172515 & SV ObjRelClauseNoThat - 0.162228 \\
            % 8 & SV SubjRelClause - 0.259019 & RA ObjRelClauseThat - 0.143430 & RA Simple - 0.124506 \\
            % 9 & SV ObjRelClauseThat - 0.211051 & SV Simple - 0.028726 & SV SubjRelClause - 0.068165 \\
            % 10 & SV ObjRelClauseNoThat - 0.179448 & SV SentComp - 0.015107 & RA SentComp - 0.017545 \\
		\hline
\end{tabular}
} % resize box
\captionof{table}{Top contributors (tasks) to top few (of 10) PCs for Transformer-XL's noun performance as detailed in Section 4.1. Cells contain the task name followed by their (absolute) component value in the eigenvector. } \label{tab:pca_contributions_txl}
\end{table*}

% pca table - contributions - bert

\begin{table*}[h!]
\centering
\resizebox{0.8\paperwidth}{!}{%
\begin{tabular}{|c||l|l|l|}
		\hline
		    Contributor & PC 1 & PC 2 & PC 3 \\
		    \hline\hline
            1 & RA ObjRelClauseNoThat - 0.456096 & SV ObjRelClauseThat - 0.577452 & SV SentComp - 0.686402 \\
            2 & RA ObjRelClauseThat - 0.444272 & SV ObjRelClauseNoThat - 0.576572 & SV Simple - 0.499513 \\
            3 & RA Simple - 0.383953 & SV PP - 0.353213 & SV SubjRelClause - 0.346437 \\
            4 & RA SentComp - 0.383866 & RA Simple - 0.288091 & SV PP - 0.248977 \\
            % 5 & SV PP - 0.297571 & RA SentComp - 0.270024 & RA ObjRelClauseThat - 0.170641 \\
            % 6 & SV Simple - 0.264730 & SV SubjRelClause - 0.166259 & RA SentComp - 0.160360 \\
            % 7 & SV ObjRelClauseNoThat - 0.255347 & SV Simple - 0.106459 & RA ObjRelClauseNoThat - 0.154101 \\
            % 8 & SV ObjRelClauseThat - 0.251706 & RA ObjRelClauseThat - 0.089542 & RA Simple - 0.118348 \\
            % 9 & SV SubjRelClause - 0.112125 & RA ObjRelClauseNoThat - 0.074722 & SV ObjRelClauseNoThat - 0.067797 \\
            % 10 & SV SentComp - 0.008082 & SV SentComp - 0.029479 & SV ObjRelClauseThat - 0.012170 \\
		\hline
\end{tabular}}
\captionof{table}{Top contributors (tasks) to top few (of 10) PCs for BERT's noun performance as detailed in Section 4.1. Cells contain the task name followed by their (absolute) component value in the eigenvector. } \label{tab:pca_contributions_bert}
\end{table*}

% pca table - contributions - gpt-2

\begin{table*}[h!]
\centering
\resizebox{0.8\paperwidth}{!}{%
\begin{tabular}{|c||l|l|l|}
		\hline
		    Contributor & PC 1 & PC 2 & PC 3 \\
		    \hline\hline
            1 & RA ObjRelClauseNoThat - 0.454492 & SV ObjRelClauseNoThat - 0.477923 & SV SentComp - 0.549969 \\
            2 & RA Simple - 0.447148 & SV SubjRelClause - 0.444648 & SV Simple - 0.525072 \\
            3 & RA SentComp - 0.441366 & SV ObjRelClauseThat - 0.426385 & SV ObjRelClauseThat - 0.478782 \\
            4 & RA ObjRelClauseThat - 0.425359 & SV SentComp - 0.385486 & SV ObjRelClauseNoThat - 0.362165 \\
            % 5 & SV PP - 0.368700 & SV Simple - 0.336849 & RA ObjRelClauseThat - 0.164547 \\
            % 6 & SV Simple - 0.246307 & RA ObjRelClauseThat - 0.206266 & SV SubjRelClause - 0.140695 \\
            % 7 & SV SubjRelClause - 0.092998 & RA Simple - 0.175984 & RA ObjRelClauseNoThat - 0.095144 \\
            % 8 & SV ObjRelClauseThat - 0.090496 & SV PP - 0.167022 & SV PP - 0.073769 \\
            % 9 & SV SentComp - 0.046917 & RA SentComp - 0.156591 & RA SentComp - 0.008498 \\
            % 10 & SV ObjRelClauseNoThat - 0.045984 & RA ObjRelClauseNoThat - 0.063824 & RA Simple - 0.000487 \\
		\hline
\end{tabular}}
\captionof{table}{Top contributors (tasks) to top few (of 10) PCs for GPT-2's noun performance as detailed in Section 4.1. Cells contain the task name followed by their (absolute) component value in the eigenvector. } \label{tab:pca_contributions_bert}
\end{table*}

% figure 1 (pairplots for bert/gpt-2)
\clearpage

\clearpage

% figure 3 (frequency plot for sv tasks)

\clearpage

% frequency plot for all tasks

%\bibliographystyle{acl_natbib}
%\bibliography{}

\end{document}